\documentclass[conference]{IEEEtran}
\IEEEoverridecommandlockouts
\usepackage{cite}
\usepackage{amsmath,amssymb,amsfonts}
\usepackage{algorithmic}
\usepackage{graphicx}
\usepackage{textcomp}
\usepackage{xcolor}
\def\BibTeX{{\rm B\kern-.05em{\sc i\kern-.025em b}\kern-.08em
    T\kern-.1667em\lower.7ex\hbox{E}\kern-.125emX}}
\newcommand{\ie}[1]{\textit{i.e.,}}
\newcommand{\eg}[1]{\textit{e.g.,}}
\newcommand{\xhdr}[1]{\vspace{1.7mm}\noindent{{\bf #1.}}}

\begin{document}

\title{Vox-UDA: Voxel-wise Unsupervised Domain Adaptation for Cryo-Electron Subtomogram Segmentation with Denoised Pseudo Labeling}

\author{\textit{Haoran Li$^{1}$, Xingjian Li$^{2}$, Jiahua Shi$^{3}$, Huaming Chen$^{4}$, Bo Du$^{5}$, Daisuke Kihara$^{6}$},\\ \textit{Johan Barthelemy$^{7}$, Jun Shen$^{1}$$^{*}$ and Min Xu$^{2}$$^{*}$}\\
$^{1}$School of Computing and Information Technology, University of Wollongong, Australia\\
$^{2}$Ray and Stephanie Lane Computational Biology Department, Carnegie Mellon University, USA\\
$^{3}$Centre for Nutrition and Food Sciences, The University of Queensland, Australia\\
$^{4}$School of Electrical and Information Engineering, University of Sydney, Australia\\
$^{5}$Department of Business Strategy and Innovation, Griffith University, Australia\\
$^{6}$Department of Biological Sciences, Purdue University, USA\\
$^{7}$NVIDIA, USA\\
\thanks{$^{*}$Corresponding author.}
}

\maketitle

\begin{abstract}\label{sec:intro}
Cryo-Electron Tomography (cryo-ET) is a 3D imaging technology facilitating the study of macromolecular structures at near-atomic resolution. Recent volumetric segmentation approaches on cryo-ET images have drawn widespread interest in biological sector. However, existing methods heavily rely on manually labeled data, which requires highly professional skills, thereby hindering the adoption of fully-supervised approaches for cryo-ET images. Some unsupervised domain adaptation (UDA) approaches have been designed to enhance the segmentation network performance using unlabeled data. However, applying these methods directly to cryo-ET images segmentation tasks remains challenging due to two main issues: 1) the source data, usually obtained through simulation, contain a certain level of noise, while the target data, directly collected from raw-data from real-world scenario, have unpredictable noise levels. 2) the source data used for training typically consists of known macromoleculars, while the target domain data are often unknown, causing the model's segmenter to be biased towards these known macromolecules, leading to a domain shift problem. To address these challenges, in this work, we introduce the first voxel-wise unsupervised domain adaptation approach, termed Vox-UDA, specifically for cryo-ET subtomogram segmentation. Vox-UDA incorporates a noise generation module to simulate target-like noises in the source dataset for cross-noise level adaptation. Additionally, we propose a denoised pseudo-labeling strategy based on improved Bilateral Filter to alleviate the domain shift problem. Experimental results on both simulated and real cryo-ET subtomogram datasets demonstrate the superiority of our proposed approach compared to state-of-the-art UDA methods.
\end{abstract}


\section{Introduction}
Cryo-Electron Tomography (cryo-ET) is one cutting-edge imaging technique which enables three-dimensional views of biological samples in a native frozen-hydrated state~\cite{oikonomou2017cellular}. This automatic electron tomography technique allows biologists to capture high-resolution structures of macromolecular complexes~\cite{wan2016cryo}, which plays an important role in the field of drug discovery and disease treatment. Inspired by the development of deep learning research in recent years, some efforts have been made in cryo-ET image analysis, especially for the subtomogram segmentation task~\cite{zhu2021weakly,zhou2021one,heebner2022deep,zhu2022unsupervised}. Subtomogram segmentation is a 3D segmentation task which aims to mine the meaningful information of the target macromolecular on the voxel-level. However, existing methods~\cite{zhou2021one,heebner2022deep,li2022domain,naylor2018segmentation} heavily rely on manual annotations which are highly subjective and resource-intensive.

\begin{figure*}[!t]
    \centering
    \includegraphics[width=1\linewidth]{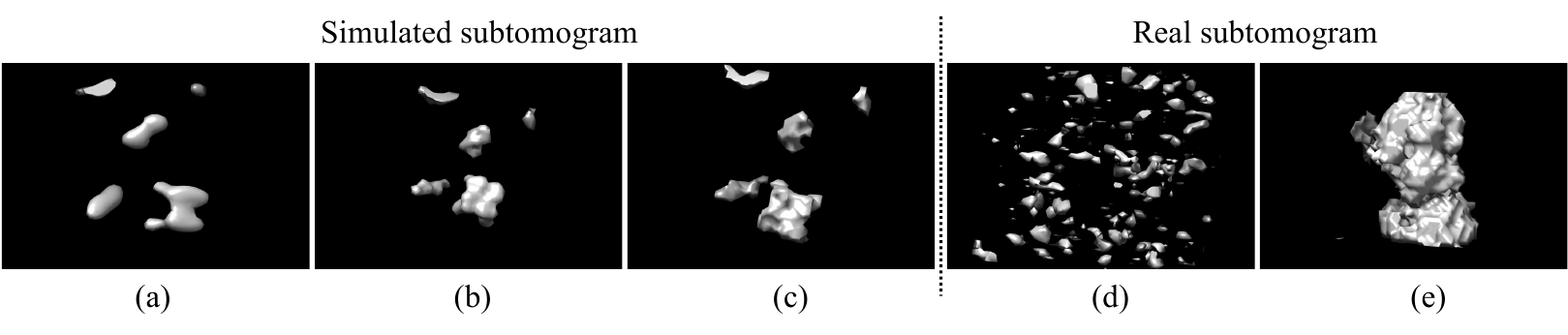}
    \caption{Some examples of the subtomograms and their corresponding segmentation masks. This figure shows: (a) simulated 3D cryo-ET subtomogram; (b) grey-scale ground truth segmentation mask; (c) binary segmentation mask after pre-processd (b), we set a threshold (300 in this paper) to turn the grey-scale mask into a binary one; (d) and (e) are real 3D cryo-ET subtomogram and its binary mask, respectively.}
    \label{fig:data}
\end{figure*}
To tackle the challenges for data annotation, the classical unsupervised domain adaptation (UDA) method involves transferring the knowledge from labeled source domains to unlabelled target domains. Ganin et al.~\cite{ganin2016domain} proposed the first UDA approach through adversarial learning, which has become the most commonly used framework for UDA tasks~\cite{van2023unpaired,zhang2023spectral}. Some other works~\cite{liu2020pdam,zhao2022uda,cicek2020disentangled} have proposed generation-based approaches which synthesizes target-like images from the source ones, and applies supervised learning using the synthesized data with their original groundtruth mask. However, these approaches were primarily designed for 2D images, and can not perform well for 3D tasks. Some recent approaches have explored UDA on 3D images~\cite{shin2023sdc,xu2023asc,xian2023unsupervised}, however, all those volumetric UDA approaches firstly cut 3D input into 2D slices for the network input, leading to the loss of spatial information. 

In this paper, we introduce one UDA approach using the large simulated macromolecular data~\cite{eisenstein2019improved,hagen2017implementation} as the source domain dataset and the real dataset as the target domain dataset. With the development of data simulation techniques, the acquisition of cryo-ET subtomogram data is no longer limited to traditional biological methods. Given the structure of macromolecules, existing generative methods~\cite{PolNet2024,harar2023faket} can directly produce realistic synthetic datasets with corresponding voxel-level segmentation masks, which can be seen as a zero-cost alternative compared to traditional methods which requires high-end equipment and enormous human expertise. Nevertheless, the significant disparities between two domains bring new challenges for the UDA task. Firstly, the simulated data is generated through fixed parameters, yielding a fixed value of the noise in each subtomogram (often 0.03 dB or 0.05 dB), while the noise rate is unpredictable in the real dataset. Some examples of the subtomograms are shown in Fig.~\ref{fig:data}. Secondly, although subtomogram segmentation is a binary segmentation task, the simulated subtomograms and the real ones often may not share the same molecular categories, which will cause the segmentation network biased to the simulated ones and lead to the domain shift problem.

To address the challenges aforementioned, we propose a voxel-wise UDA framework, termed Vox-UDA, for cryo-ET subtomogram segmentation. Vox-UDA consists of a noise generation module (NGM) and a denoised pseudo-labeling (DPL) strategy. NGM generates Gaussian noise from a subset of the target dataset and applies it to the source samples to create a target-like noisy phenomenon. Meanwhile, DPL improves the existing bilateral filter, making it more suitable for 3D grayscale images through modifying the pixel difference of one Gaussian kernel to gradient difference. While denoising, DPL preserves edge information as much as possible to obtain undistorted pseudo-labels. These pseudo-labels provide additional supervision signals to address the domain shift problem, thereby enhancing the model's performance on the target data.

In a nutshell, our contributions are as follows:
\begin{itemize}
    \item[$\bullet$] To the best of our knowledge, herein we are the first to esablish a paradigm for voxel-wise UDA segmentation in cryo-ET images (termed Vox-UDA). Our approach eliminates the reliance on large amounts of labeled real data by transferring knowledge learned from zero-cost simulated data to the real ones, enabling segmentation on real cryo-ET subtomograms.
    \item[$\bullet$] Our Vox-UDA includes a noise generation module (NGM) and a denoised pseudo-labeling (DPL) strategy to enable the simulation of target-like noisy phenomenon, and it provide additional supervision signals to address the domain shift problem.
    \item[$\bullet$] We propose an improved bilateral filter that, by being sensitive to the changes in gradients, preserves edge information as much as possible while eliminating noises in order to obtain high-quality pseudo-labels.
    \item[$\bullet$] The extensive experimental results demonstrate the superiority of Vox-UDA method over state-of-the-art UDA methods on subtomogram segmentation. Besides, our method even outperforms fully supervised methods on some metrics.
\end{itemize}

The rest of the paper is organized as follows. A brief literature review is presented in Section~\ref{sec:related}. We provide the details of our proposed Vox-UDA in Section~\ref{sec:method}. Experimental results and visualizations are shown in Section~\ref{sec:exp}, followed be the conclusion in Section~\ref{sec:conclu}.

\begin{figure*}[!t]
    \centering
    \includegraphics[width=1\linewidth]{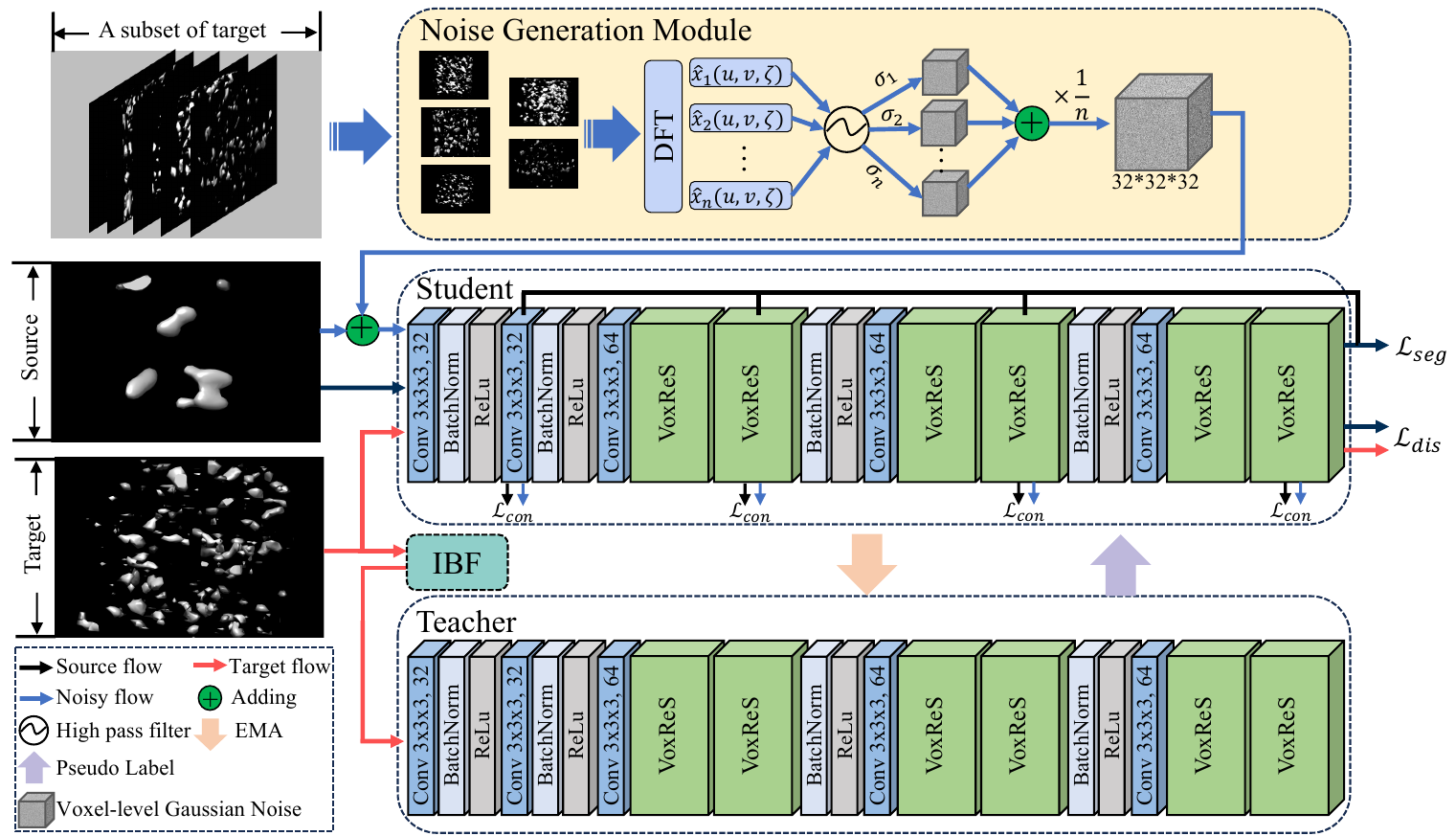}
    \caption{Overview of our proposed Vox-UDA framework. IBF denotes the improved Bilateral Filter, which is detailed in Fig~\ref{fig:DPL}. We use different colors to represent different flows. Best viewed in color.}
    \label{fig:framework}
\end{figure*}

\section{Related Work}\label{sec:related}
\subsection{Unsupervised Domain Adaptation for Vision Tasks}
Under the unsupervised domain adaptation (UDA) settings, there are two types of dataset being used for training: the source domain dataset which is fully labelled, and the target domain dataset, which is unlabelled. The first UDA approach is proposed by~\cite{ganin2016domain}, which aims to transfer the model trained on source data to target data without introducing additional annotations through adversarial learning. Since UDA greatly expands the model's generalization ability, the model can be adapted to new domains without requiring labeled data in the target domain and is introduced into various tasks, \eg, classification~\cite{sharma2021instance,xiao2021dynamic,zhang2022spectral,yang2023tvt}, object detection~\cite{guan2021uncertainty,yu2022sc,mattolin2023confmix,yoo2022unsupervised} and semantic segmentation~\cite{dong2021and,lee2021unsupervised,zhu2023unsupervised,zhao2024unsupervised}. As cryo-ET subtomogram segmentation is a segmentation task, in this paper, we mainly focus on the UDA approaches applied in semantic segmentation tasks. For segmentation, the UDA methods aim to eliminate the cross-domain discrepancies through the content at both the feature- and pixel-level. Zou \textit{et al.} proposes a class-aware UDA approach based on self-training to handle the class-imbalance problem~\cite{zou2018unsupervised}. Zheng \textit{et al.} designs a dual-path framework, which fuses equirectangular projections and tangent projection for Panoramic Semantic Segmentation~\cite{zheng2023both}. Additionally, UDA has achieved excellent results in medical image segmentation~\cite{zhang2023st,xie2022unsupervised,li2022domain}. Ji \textit{et al.} introduces an attention-based method, which learns the hierarchical consistencies and transfer more discriminative information between the source and target domain~\cite{ji2023unsupervised}. However, although these methods achieves great performance in UDA segmentation, they are primarily designed for 2D images, which are not suitable for cryo-ET subtomogram segmentation as tomographies are often volumetric images. To handle this UDA challenge in 3D segmentation tasks, Shin et al.~\cite{shin2023sdc} proposed a cross-modality translation method to generate synthetic 3D target volumes from source 2D scans. Xu et al.~\cite{xu2023asc} applied a fast Fourier transform to convert input 2D slices into frequency domain. A consistency loss was utilised to simultaneously constrain both the feature domain and frequency domain to achieve UDA. As discussed in Sec~\ref{sec:intro}, all the exiting 3D UDA approaches convert images into 2D UDA tasks by slicing, rather than directly transferring in the three-dimensional voxel space, which will lead to information loss.

\subsection{Cryo-Electron Tomography}
Cryo-electron tomography (cryo-ET) integrates cryogenic specimen preparation, electron microscopy for data acquisition, and tomographic reconstruction for 3D visualization~\cite{koning2010chapter}. This technique allows to capture structural information of macromolecules in an ultra-low temperature environment, which holds significant importance in the fields of biology and medicine. A cryo-ET subtomogram is a small cubic sub-volume extracted from a tomogram, normally only a single macromolecular complex is contained in each subtomogram. Inspired by recent advancements in deep learning, their applications on cryo-ET have drawn widespread interests with their potential to aid in the corresponding cryo-ET tasks, \eg, subtomogram alignment~\cite{zeng2020gum,de2024deep}, subtomogram classification~\cite{wan2024stopgap,du2021active} and subtomogram segmentation~\cite{zhu2022unsupervised,nguyen2022finding,siggel2024colabseg}. However, deep learning methods rely on large amounts of data annotation, which is particularly challenging for cryo-ET images. Bandyopadhyay \textit{et al.} proposes a domain randomization-based approach to enhance the generalization performance of the model in subtomogram classification~\cite{bandyopadhyay2022cryo}. Zhu \textit{et al.} proposed a weakly supervised approach which used only 2D-level annotation for voxel segmentation to alleviate the burden of annotation~\cite{zhu2021weakly}. In this paper, we will propose an UDA approach for subtomogram segmentation, which aims at utilizing a large amount of cost-free annotated simulated data for knowledge transfer, enabling the segmentation network to generalize on real cryo-ET subtomograms.

\section{Method}\label{sec:method}

Our proposed framework is based on VoxResNet~\cite{chen2018voxresnet}, a state-of-the-art method designed for fully-supervised voxel-level segmentation. As can be seen from Fig.~\ref{fig:framework}, VoxResNet takes the combination of the outputs from the second convolution layer, the second VoxReS module, the fourth VoxReS module, and the last VoxReS module as its final output.

Given a source domain dataset $\mathcal{S} = \left\{x^{s}_{i}, y^{s}_{i} \right\}^N_{i=1}$ and a target domain dataset $\mathcal{T} = \left\{x^{t}_{j}\right\}^M_{j=1}$, where $x_i$ represents the input 3D subtomogram and $y_i$ denotes the 3D groundtruth mask, we aim to train a voxel segmentation network for the target domain only using groundtruth supervision signals from the source domain. Fig.~\ref{fig:framework} illustrates the details of the proposed Vox-UDA. As can be seen from the figure, Vox-UDA takes $x^{s}_{i}$, $x^{t}_{i}$ and a subset of $\mathcal{T}$ as input. This subset $\mathcal{T}_{N_{sampled}}$ is randomly sampled from $\mathcal{T}$, which contains $N_{sampled}$ samples. $\mathcal{T}_{N_{sampled}}$ is then sent to the noise generation module (NGM) to obtain the target-like voxel-wise Gaussian noise $\epsilon \sim \mathcal{N}(0, \sigma_{t}^{2}\mathbf{I})$. Further,  $\epsilon$ is introduced to the source input $x^{s}_{i}$ to produce updated input $x^{s'}_{i}$. $x^{s}_{i}$, $x^{s'}_{i}$ and $x^{t}_{j}$ are all passed to the student network to acquire segmentation loss $\mathcal{L}_{seg}$, consistency loss $\mathcal{L}_{con}$ and discriminator loss $\mathcal{L}_{dis}$ for optimization. Following~\cite{li2022domain,xu2023asc}, we set the same weight for different losses. Hence, the overall loss can be rewritten as
\begin{equation}
    \mathcal{L} = \mathcal{L}_{seg} + \mathcal{L}_{con} + \mathcal{L}_{dis}.
\end{equation}
Furthermore, to handle the domain shift problem, we design a denoised pseudo-labeling strategy. $x^{t}_{j}$ is sent to the improved Bilateral Filter (IBF) to eliminate its noise and then sent to the teacher network to obtain the pseudo-label, which is then used to tune the student network for better performance. Noted that the threshold $\eta$ used for pseudo-label selection is set to $0.85$ and the teacher network is updated via exponential moving average (EMA).

\subsection{Noise Generation Module}\label{sec:NGM}
Inspired by recent approaches, such as ~\cite{ho2020denoising,nichol2021improved,lehtinen2018noise2noise}, using the segmentation network as the denoiser for noise elimination in 2D space, we extend this insight to three-dimensional space and propose a new noise generation module. Given a sample $x^{t}_{n}$ from the input $\mathcal{T}_{N_{sampled}}$, we first apply Discrete Fourier transform (DFT) to obtain its frequency information
\begin{equation}
    \hat{x}_n(u,v,\zeta) = \xi[x^{t}],
\end{equation}
where $u$, $v$ and $\zeta$ represent the spatial frequencies of the Fourier transform, and $\xi$ denotes the Discrete Fourier transform. In the frequency domain, low-frequency information corresponds to the textural details of the target object, while large amounts of noise with little edge information about the object are usually encompassed into the high-frequency information. To obtain the noise encompassed in the high-frequency information, $x_n(u,v,\zeta)$ is then passed to a high-pass filter to eliminate the textural details contained in low-frequency information
\begin{equation}
    \hat{x}_n'(u,v,\zeta) = H_{high}(u,v,\zeta)\hat{x}_n(u,v,\zeta), 
\end{equation}
where $H_{high}(u,v,\zeta)$ denotes the high-pass filter. The filter rate is set to $24.4\%$, which means only $24.4\%$ remains while the rest of them are filtered (see Sec.~\ref{sec:ablation} for detailed discussions). Inverse Discrete Fourier transform (iDFT) is further applied to recover voxel-level information from the filtered frequency domain $x_n'(u,v,\zeta)$:
\begin{equation}
    x^{t'}_n = \xi^{-1}[\hat{x}_n'],
\end{equation}
\begin{figure}[!t]
    \centering
    \includegraphics[width=1\linewidth]{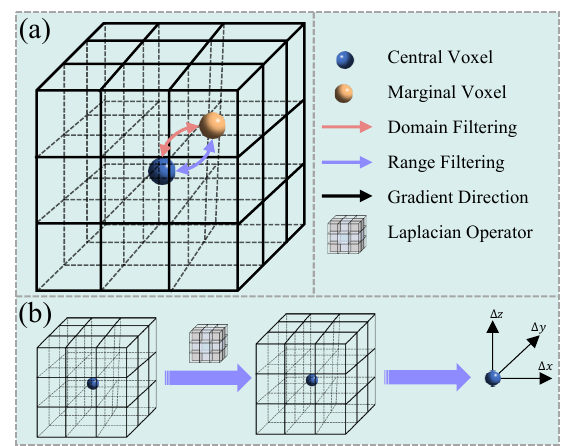}
    \caption{Proposed improved Bilateral Filter. (a) Both domain filtering and range filtering are applied to an sub-figure extracted from the input target subtomogram with size $3 \times 3 \times 3$. (b) Deploying Laplace transform to obtain the gradient changes used in range filtering.}
    \label{fig:DPL}
\end{figure}
where $\xi^{-1}[\cdot]$ denotes the Inverse Discrete Fourier Transform. As discussed in Sec~\ref{sec:intro}, the noise level of each input from the target domain is unpredictable, hence instead of using the noise from single $x^{t}_{n}$, we calculate the average noise level $\overline{x}^{t'}_n$ from the whole subset. 
Since deep learning models are more sensitive to noise that conforms to a probability distribution~\cite{lehtinen2018noise2noise}, we set the Gaussian Noise as the input noise for noise generation (we also compare other types of noise, ablation studies are provided in Sec~\ref{sec:NGP}). 
Therefore, instead of directly introducing $\overline{x}^{t'}_n$ to the source input $x^{s}_{i}$, we only take its variance $\sigma^{2}_{t}$ and generate a Gaussian noise based on 
\begin{equation}
    \epsilon = \mathcal{N}(0, \sigma_{t}^2 \mathbf{I}),
\end{equation}
where $\mathcal{N}(0, \sigma_{t}^2 \mathbf{I})$ denotes a random generated Gaussian noise with expectation equals to $0$, and variance equals to $\sigma_{t}^2$. And the updated source input is obtained through
\begin{equation}
    x^{s'}_{i} = x^{s}_{i} + \epsilon.
\end{equation}

\begin{table*}[!t]
    \centering
    \caption{Comparison of experimental results on UDA cryo-ET subtomogram segmentation between Vox-UDA and previous non-voxel-level UDA approaches. Best results are in bold font.}
    \label{tab:source1}
    \begin{tabular}{l|c|c|c|c|c|c|c|c}
        \hline
         & \multicolumn{8}{c}{Source macromoleculars: 1bxn, 1f1b and 1yg6}\\
         \hline
         Method & mIoU & mIoU$_{ribo}$ & mIoU$_{26S}$ & mIoU$_{TRiC}$ & Dice & Dice$_{ribo}$ & Dice$_{26S}$ & Dice$_{TRiC}$\\
         \hline
         w/o adaptation & 9.7 & 11.1 & 2.6 & 2.6 & 17.4 & 19.9 & 5.1 & 5.0\\
         Fully Supervised & 46.0 & 49.5 & 20.0 & 34.6 & 61.6 & 65.7 & 30.0 & 48.3\\
         \hline
         DANN~\cite{ganin2016domain} & 38.4 & 43.0 & 6.5 & 11.7 & 53.0 & 59.1 & 11.2 & 16.8\\
         PDAM~\cite{liu2020pdam} & 39.8 & 43.3 & 13.8 & 22.6 & 55.1 & 59.6 & 21.5 & 31.9\\
         ASC~\cite{xu2023asc} & 40.4 & 43.4 & 19.2 & 23.3 & 55.8 & 59.7 & 28.7 & 32.7\\
         LE-UDA~\cite{zhao2022uda} & 41.5 & 44.7 & 18.4 & 23.9 & 56.8 & 61.0 & 28.4 & 32.6\\
         \hline
         \hline
         \textbf{Vox-UDA(w NGM)}& 48.5 & 50.6 & \textbf{32.2} & 38.6 & 64.4 & 66.8 & \textbf{47.1} & 51.5\\
         \textbf{Vox-UDA(w BF)}& 49.1 & 50.4 & 30.5 & 39.2 & 64.5 & 67.1 & 46.9 & 50.7\\
         \textbf{Vox-UDA(w IBF)}& \textbf{50.3} & \textbf{53.8} & 28.8 & \textbf{41.3 } & \textbf{65.9} & \textbf{68.5} & 44.0 & \textbf{52.8}\\
         \hline
    \end{tabular}
\end{table*}

\begin{figure*}[!t]
    \centering
    \includegraphics[width=1\linewidth]{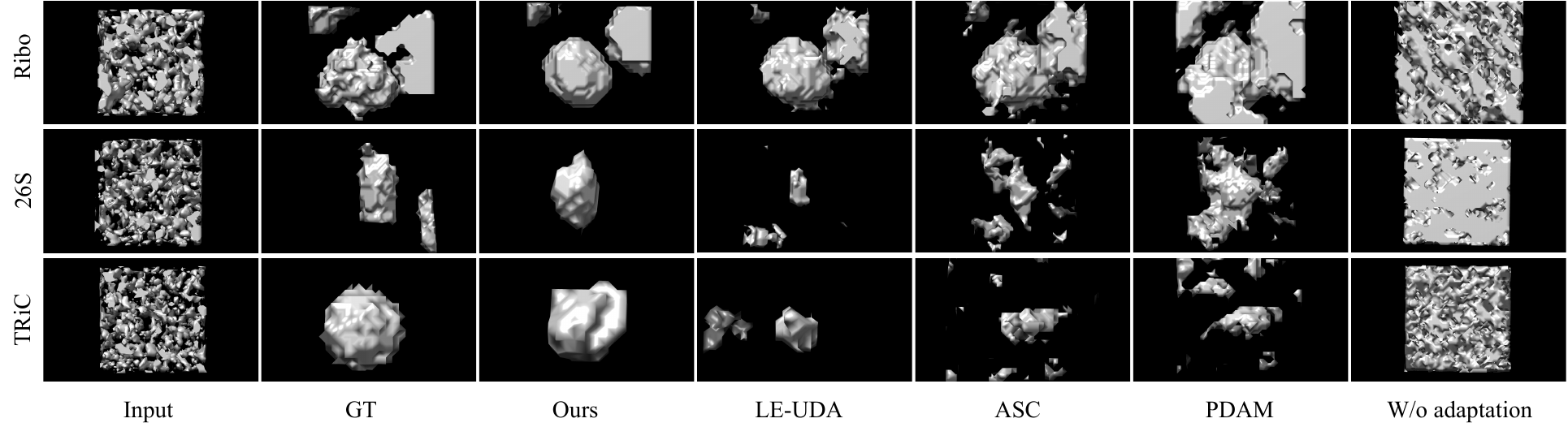}
    \caption{Visualization of subtomogram segmentation results using 1bxn, 1f1b and 1yg6 as the source datasets. We use UCSF Chimera~\cite{pettersen2004ucsf} for 3D cryo-ET visualization.}
    \label{fig:vis}
\end{figure*}

$x^{s}_{i}$ and $x^{s'}_{i}$ are both sent to the student network to obtain the consistency loss $\mathcal{L}_{con}$. Following VoxResNet, we also take the output feature embeddings from the same layers as the input of the loss function
\begin{equation}
\begin{aligned}\label{eq:consis}
    \mathcal{L}_{con} = \lambda_{1}\mathcal{L}_{BN}(f_c, f'_c) + \lambda_{2}\mathcal{L}_{BN}(f_{v2}, f'_{v2}) + \lambda_{3}\mathcal{L}_{BN}(f_{v4}, f'_{v4})&\\ + \lambda_{4}\mathcal{L}_{BN}(f_{v6}, f'_{v6}),
\end{aligned}
\end{equation}
where $\mathcal{L}_{BN}$ denotes the cosine similarity loss and $\lambda_{n}$ denotes the weights to control the relative importance among different consistency losses. Although NGM is introduced to simulate target-like noises, it is impossible to create a noise environment in the source domain that is entirely the same as the target domain. On the other hand, while the shallower layers of the decoder containing more textural information, the deeper layers contain more edge information~\cite{chen2017deeplab}. To overcome these constraints, our solution is, instead of an equal superposition, we assign different weights to the different layers to control the weighting of texture and edge consistency losses (see detail discussions in Sec~\ref{sec:ablation}).

\subsection{Denoised Pseudo-Labeling}\label{sec:DPL}
Although the NGM can narrow noise level gaps between two domains, as aforementioned, the segmentation network is still biased to the source data due to the domain shift problem. Therefore, we provide an extra supervision signal for optimization through pseudo-labeling. However, due to the noise level being unknown in the target domain and also that such noise may lead to distorted pseudo-labels further harming the performance of the model, we propose a denoised pseudo-labeling strategy instead. Unlike the existing pseudo-labeling method whereby adding an extra training step, we use the student-teacher structure~\cite{sohn2020fixmatch}. Before $x^{t}_{j}$ is sent to the teacher network to obtain the pseudo-label, we first perform denoising on $x^{t}_{j}$. We designed three different denoising methods: $1)$ directly using NGM for denoising, $2)$ using Bilateral Filter for denoising, and $3)$ using our designed improved Bilateral Filter (IBF) for noise reduction.

\begin{table*}[!t]
    \centering
    \caption{Experimental results on UDA cryo-ET subtomogram segmentation under different dataset settings. Different from the results reported in Table~\ref{tab:source1}, we set 2byu, 2h12 and 21db as the source datasets.}
    \label{tab:source2}
    \begin{tabular}{l|c|c|c|c|c|c|c|c}
        \hline
         & \multicolumn{8}{c}{Source macromoleculars: 2byu, 2h12 and 21db}\\
         \hline
         Method & mIoU & mIoU$_{ribo}$ & mIoU$_{26S}$ & mIoU$_{TRiC}$ & Dice & Dice$_{ribo}$ & Dice$_{26S}$ & Dice$_{TRiC}$\\
         \hline
         w/o adaptation & 12.7 & 14.2 & 3.7 & 3.0 & 22.2 & 24.7 & 7.2 & 5.8\\
         Fully Supervised & 46.0 & 49.5 & 20.0 & 34.6 & 61.6 & 65.7 & 30.0 & \textbf{48.3}\\
         \hline
         DANN~\cite{ganin2016domain} & 31.9 & 36.1 & 3.8 & 6.2 & 45.9 & 51.9 & 6.4 & 8.7\\
         PDAM~\cite{liu2020pdam} & 39.1 & 43.1 & 10.9 & 15.7 & 54.1 & 59.4 & 17.7 & 24.0\\
         ASC~\cite{xu2023asc} & 41.7 & 45.2 & 24.7 & 13.3 & 56.9 & 61.2 & 38.1 & 19.8\\
         LE-UDA~\cite{zhao2022uda} & 43.1 & 46.4 & 21.9 & 22.3 & 58.6 & 62.6 & 33.8 & 32.4\\
         \hline
         \hline
         \textbf{Vox-UDA(w NGM)}& 47.5 & 50.1 & 27.5 & 34.7 & 63.2 & 66.3 & 41.0 & 46.8\\
         \textbf{Vox-UDA(w BF)}& 48.0& 50.3& 27.8& 34.4& 63.8& 66.7& 39.9& 47.0\\
         \textbf{Vox-UDA(w IBF)}& \textbf{49.5}& \textbf{52.4}& \textbf{28.3}& \textbf{35.1}& \textbf{65.2}& \textbf{68.9}& \textbf{41.3}& 47.7\\
         \hline
    \end{tabular}
\end{table*}

\begin{figure*}[!t]
    \centering
    \includegraphics[width=1\linewidth]{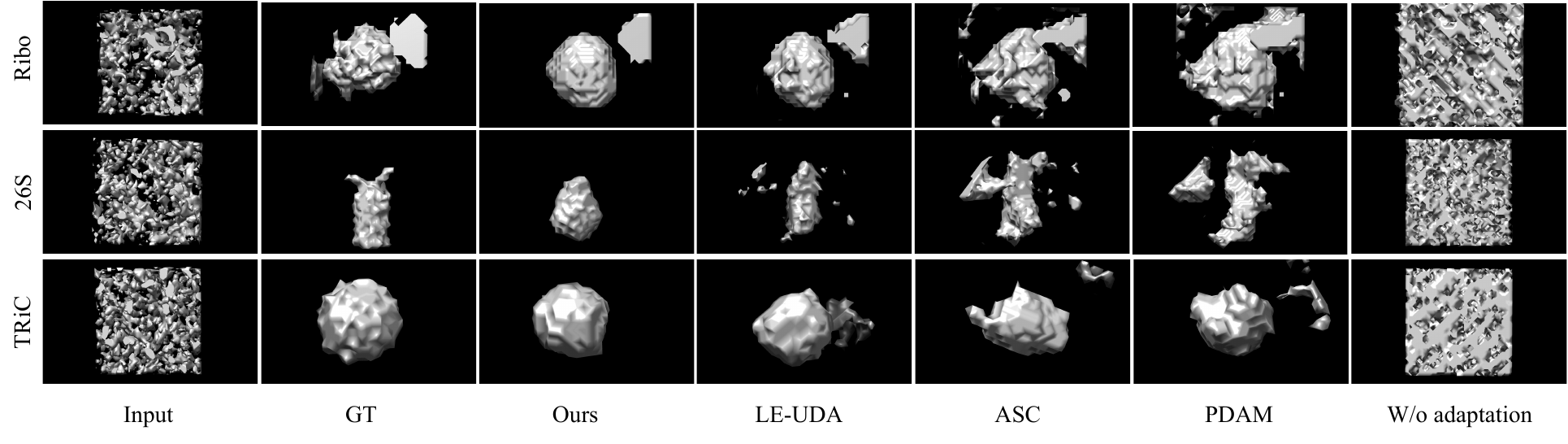}
    \caption{Additonal visualization of subtomogram segmentation results using 2byu, 2h12 and 21db as source dataset.}
    \label{fig:vis2}
\end{figure*}

\xhdr{NGM Denoising} The $x^{t}_{j}$ is directly sent to the NGM to obtain its noise $x^{t'}_{j}$. Hence, the denoised image can be represented as $\widetilde{x}^{t}_{j}=\left(x^{t}_{j} - x^{t'}_j\right)$.

\xhdr{Bilateral Filter Denoising} Although noise can be partially removed through frequency domain analysis, some edge information will also be eliminated, leading to distortion of the pseudo-labels. Therefore, we further deploy a non-linear approach, Bilateral Filter~\cite{elad2002origin}, as the denoiser instead of the NGM. Bilateral filter (BF) consists of a domain Gaussian kernel and a range Gaussian kernel, the former is used to eliminate the noises, and the later is to retain edge information as much as possible during filtering. BF uses a sliding window, extracting a $3 \times 3 \times 3$ sub-figure for filtering operations each time. Given the central voxel $v_p$ and the rest voxels $v_q, q \in V$ of the sub-figure, the updated voxel can be represented as
\begin{equation}\label{eq:BF}
    v^{'}_{q} = \text{BF}(v_{q}) = \frac{\sum_{q \in V} G_{\sigma_{d}}(||p-q||)G_{\sigma_{r}}(v_{q} - v_{p})\times v_{q}}{\sum_{q \in V} G_{\sigma_{d}}(||p-q||)G_{\sigma_{r}}(v_{q} - v_{p})},
\end{equation}
where $||\cdot||$ denotes the Euclidean distance~\cite{wang2005euclidean}, $\sigma_{d}$ and $\sigma_{r}$ denote the domain hyperparameter and range hyperparameter, and $G_{\sigma}$ denotes the Gaussian kernel
\begin{equation}
    G_{\sigma}(x) = \frac{e^{-\frac{x^{2}}{2\sigma^{2}}}}{2\pi \sigma}.
\end{equation}
Hence, the denoised image can be represented as $\widetilde{x}^{t}_{j} = \text{BF}(x^{t}_{j})$.

\xhdr{IBF Denoising} The key point of the Bilateral Filter is the design of using two separate Gaussian kernels for different tasks. However, the range kernel introduced for retaining edge information mainly focuses on the voxel-level color difference, which indeed can achieve satisfactory results in the RGB space, but in the grayscale space, there might be more differences in brightness, which coulc affect the effectiveness of this kernel. Therefore, we further propose an improved Bilateral Filter (IBF), which uses the gradient of each voxel instead of its value for the range kernel for edge retaining. In detail, we reflect Laplace operator~\cite{van1989nonlinear} into 3-dimension and calculate the gradient of each voxel $v_{q}$ in the $h$, $w$, and $d$ (height, width and depth) directions in a three-dimensional space. Since voxel space is discrete, the gradient of $v_{q}$ in each direction can be represented as
\begin{align}
    \frac{\partial \Delta}{\partial \Vec{v}_{q}^{h}} = \Delta(v_{q+1}^{h}, v_{q}^{w}, v_{q}^{d}) - \Delta(v_{q-1}^{h}, v_{q}^{w}, v_{q}^{d}),\\
    \frac{\partial \Delta}{\partial \Vec{v}_{q}^{w}} = \Delta(v_{q}^{h}, v_{q+1}^{w}, v_{q}^{d}) - \Delta(v_{q}^{h}, v_{q-1}^{w}, v_{q}^{d}),\\
    \frac{\partial \Delta}{\partial \Vec{v}_{q}^{d}} = \Delta(v_{q}^{h}, v_{q}^{w}, v_{q+1}^{d}) - \Delta(v_{q}^{h}, v_{q}^{w}, v_{q-1}^{d}),
\end{align}
where $x_{q}^{h}$, $x_{q}^{w}$ and $x_{q}^{d}$ denote the values of $v_{q}$ in the $h$, $w$, and $d$ directions, and $\Delta$ denotes the Laplace operator. Moreover, compared to the gradient of the central voxel $v_p$, if a voxel is belonging to the object (inside), their gradient should be similar. Otherwise, the difference between two gradient should be large. Therefore, we can replace the second filter in Eq~\ref{eq:BF} and obtain the improved Bilateral Filter (IBF):
\begin{equation}\label{eq:IBF}
    v^{'}_{q} = \text{IBF}(v_{q}) = \frac{\sum_{q \in V} G_{\sigma_{d}}(||p-q||)G_{\sigma_{r}}(\frac{\partial \Delta}{\partial \Vec{v}_{p}} - \frac{\partial \Delta}{\partial \Vec{v}_{q}})\times v_{q}}{\sum_{q \in V} G_{\sigma_{d}}(||p-q||)G_{\sigma_{r}}(\frac{\partial \Delta}{\partial \Vec{v}_{p}} - \frac{\partial \Delta}{\partial \Vec{v}_{q}})},
\end{equation}
where
\begin{equation}
    \frac{\partial \Delta}{\partial \Vec{v}_{p}} - \frac{\partial \Delta}{\partial \Vec{v}_{q}} = (\frac{\partial \Delta}{\partial \Vec{v}_{p}^{h}}, \frac{\partial \Delta}{\partial \Vec{v}_{p}^{w}}, \frac{\partial \Delta}{\partial \Vec{v}_{p}^{d}}) - (\frac{\partial \Delta}{\partial \Vec{v}_{q}^{h}}, \frac{\partial \Delta}{\partial \Vec{v}_{q}^{w}}, \frac{\partial \Delta}{\partial \Vec{v}_{q}^{d}}).
\end{equation}
Consequently, the denoised image is denoted as $\widetilde{x}^{t}_{j} = \text{IBF}(x^{t}_{j})$.

$\widetilde{x}^{t}_{j}$ is further sent to the teacher network and obtain the pseudo-label with the threshold $\eta$. The pseudo-label is further sent back to the student network as a supervision signal for the target flow.

\section{Experiments}\label{sec:exp}

\subsection{Experimental Settings}

\subsubsection{Datasets and Evaluation Metrics}
We conduct experiments on two types of datasets: simulated dataset and real dataset. 

\xhdr{Simulated Dataset} The simulated dataset used as source dataset is generated following the same generation process as ~\cite{zeng2023high}. We choose six representative macromolecule complexes in our simulated datasets and divide them into two groups as two separate source datasets (1bxn, 1f1b, and 1yg6; 2byu, 2h12, and 21db). For each macromolecule complex, we simulate it with two different noise levels, with SNR of 0.03 and 0.05, and each of them contains 500 samples. Following existing work~\cite{zeng2020gum,liao2020iteratively}, all the input subtomogram are resized to $32^{3}$. The simulated dataset contains 6,000 samples in total (3,000 samples for each source dataset). 

\xhdr{Real Dataset} The real dataset used as the target dataset is the public dataset Poly-GA~\cite{guo2018situ}, which contains 66 $26S$ subtomograms, 66 $TRiC$ subtomograms and 901 $Ribosome$ subtomograms (1,033 samples in total). Each subtomogram is also re-scaled to size $32^3$.

For evaluation, the mean intersection of union (mIoU) and dice similarity coefficient (Dice) are employed to evaluate the segmentation performance.
\subsubsection{Implementation Details}
We utilize the VoxResNet as our base architecture. The whole model is trained on a single NVIDIA A100 Tensor Core GPU with 80GB memory. For training, we choose the Adam optimizer with an initial learning rate set to 1e-3 for optimization. The model is trained for 300 epochs with batch size of 16. The learning rate is decayed by $90\%$ every 100 epochs. The hyperparameters sampled number $N_{sampled}$ and filter rate $\rho$ are empirically set to 10 and $24.4\%$, separately (see discussions in Sec~\ref{sec:ablation}). For the improved Bilateral Filter, the domain hyperparameter $\sigma_{d}$ and range hyperparameter $\sigma_{r}$ are set to $120$ and $1.2$, respectively. We use Sobel operator as the Laplacian operator.

\begin{figure}[!t]
    \centering
    \includegraphics[width=1\linewidth]{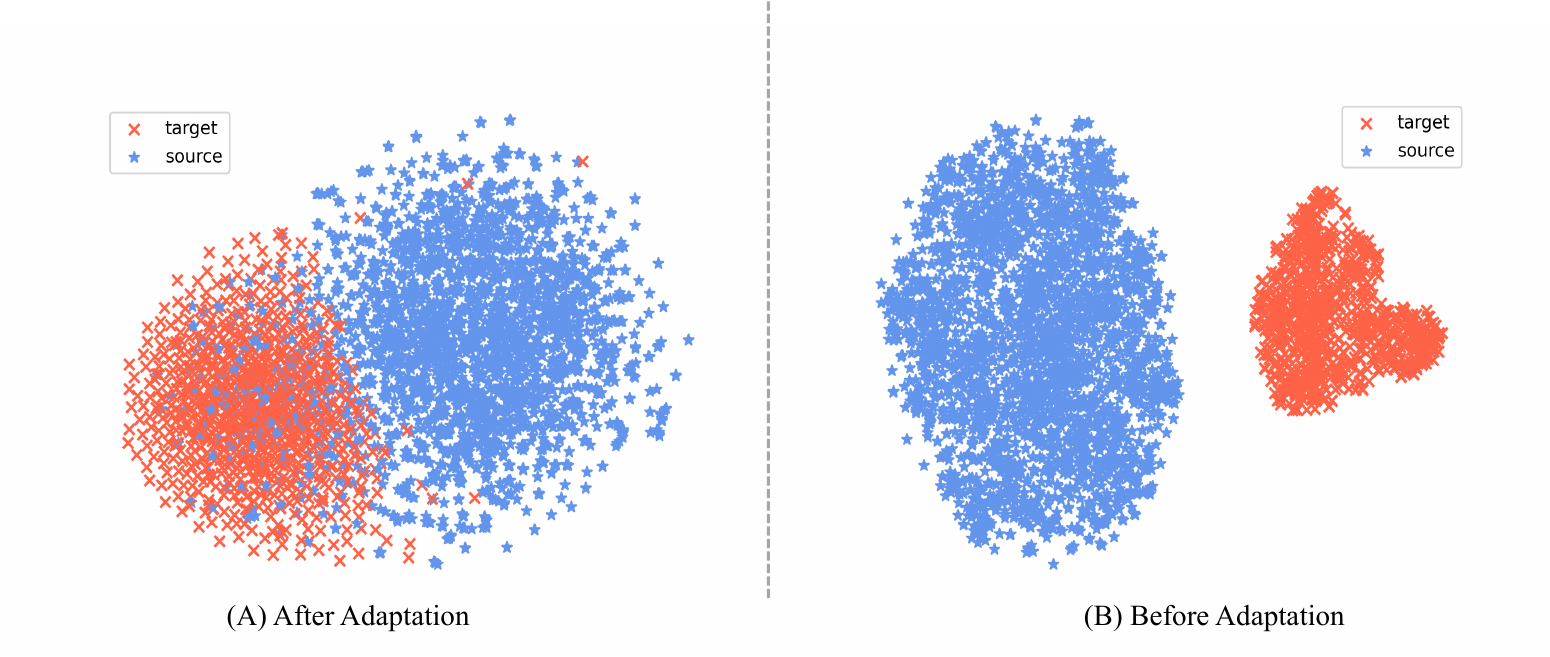}
    \caption{The t-SNE visualization of learned features. (a) After UDA. (b) Before UDA. }
    \label{fig:tSNE}
\end{figure}

\subsubsection{Baselines}
As there are no existing methods designed for voxel-level UDA, we implement several traditional and state-of-the-art UDA approaches for 2D image segmentation on our task, including single discriminator-based (DANN~\cite{ganin2016domain}) and image synthesizing based (PDAM~\cite{liu2020pdam}). And we also include two most recent approaches designed for volumetric images(ASC~\cite{xu2023asc} and LE-UDA~\cite{zhao2022uda}), which cut 3D images into 2D slices and apply UDA on 2D scenario. Following existing UDA methods~\cite{liu2020pdam,ji2023unsupervised,zhu2023unsupervised}, we also set a ``w/o adaptation'' setting and a ``Fully Supervised'' setting for comparison. The ``w/o adaptation'' setting is a original VoxResNet trained on source dataset without adaptation. The ``Fully Supervised'' setting is a VoXResNet fully supervised trained on the labelled target datasets, as the upper bound.

\subsection{Comparisons With State-of-the-arts}

We report the segmentation results on the Poly-GA dataset in Table~\ref{tab:source1} using the [1bxn, 1f1b, and 1yg6] as the source dataset. As can be observed in the table, our approach outperforms all the state-of-the-art methods. Compared with ``w/o adaptation'', DANN and PDAM indeed boost the model's performance, however, the effect is not obvious compared with our Vox-UDA (w IBF) (\ie, PDAM achieves $55.1$ in $Dice$ while Vox-UDA (w IBF) achieves $65.9$). And compared with two recent UDA methods, our proposed Vox-UDA (w IBF) still excels on target subtomogram segmentation, which leads to significant improvements in both $mIoU$ (\ie, $40.4 \rightarrow 50.3$) and $Dice$ (\ie, $56.8 \rightarrow 65.9$). We also report extra UDA setting results in Table~\ref{tab:source2} by using the other three macromoleculars [2byu, 2h12, and 21db] as source datasets, by which our proposed method still achieves state-of-the-art performance over all the comparison approaches. It is worth noting that in both tables, our Vox-UDA even surpasses the ``fully supervised'' setting on the vast majority of the metrics (\ie, ``Fully Supervised'' achieves $46.0$ in $mIoU$ in Table~\ref{tab:source2}, while our  Vox-UDA (w IBF) achieves $49.5$).

\subsubsection{Segmentation results Visualization}
Fig~\ref{fig:vis} shows the segmentation results on the Poly-GA dataset using 1bxn, 1f1b and 1yg6 as source dataset. As can be observed, due to the proposed noise generation module can simulate the target noise environment on the source data, our model's robustness to noise is significantly enhanced (\ie, compared to the segmentation results of ASC and PDAM, ours results focus more on the macromolecules rather than the surrounding noise). Compared with LE-UDA, our segmentation results exhibit better texture details due to the proposed denoised pseudo labeling strategy, which avoids the model being biased towards source data and addresses the domain shift problem. We also provide additional visualization results using 2byu, 2h12 and 21db as source dataset in Fig~\ref{fig:vis2}.

\subsubsection{Feature Visualization}
As shown in Fig~\ref{fig:tSNE}, we visualize the feature embeddings learned by the segmentation network with the commonly used t-SNE~\cite{van2008visualizing} method. Fig~\ref{fig:tSNE}(a) and Fig~\ref{fig:tSNE}(b) represent the visualization results of ``after adaptation'' and ``before adaptation'', respectively. As can be observed from the figure, the distribution of the source and target features learned from our proposed method is more consistent compared to the distribution without adaptation. This indicates that our method has achieved knowledge transfer and generalized the model to the target data.

\subsection{Ablation Study}\label{sec:ablation}
\begin{table*}[!t]
    \centering
    \caption{Ablation study on the components of our proposed model on Poly-GA dataset. NGM and PL denote noise generation module and pseudo-label, respectively. Best results are in bold font.}
    \label{tab:ablation1}
    \begin{tabular}{l|c|c|c|c|c|c|c|c}
        \hline
         & \multicolumn{8}{c}{Source macromoleculars: 1bxn, 1f1b and 1yg6}\\
         \hline
         Method & mIoU & mIoU$_{ribo}$ & mIoU$_{26S}$ & mIoU$_{TRiC}$ & Dice & Dice$_{ribo}$ & Dice$_{26S}$ & Dice$_{TRiC}$\\
         \hline
         Baseline & 31.9 & 36.1 & 3.8 & 6.2 & 45.9 & 51.9 & 6.4 & 8.7\\
         w/o NGM & 41.9 & 43.9 & 20.1 & 37.9 & 57.5 & 60.2 & 31.0 & 50.4\\
         w/o PL & 46.2 & 49.1 & 23.3 & 31.6 & 61.8 & 65.3 & 36.3 & 44.0\\
         \textbf{Vox-UDA(Ours)}& \textbf{50.3} & \textbf{53.8} & \textbf{28.8} & \textbf{41.3} & \textbf{65.9} & \textbf{68.5} & \textbf{44.0} & \textbf{52.8}\\
         \hline
    \end{tabular}
\end{table*}

\begin{table*}[!t]
    \centering
    \caption{Ablation study on the hyperparameters of our proposed model on Poly-GA dataset. N$_{sample}$, $\rho$ and $\lambda_{n}$ denote the sampled number of target data used for noise generation, the high-pass filter rate and the weights for different losses, respectively. }
    \label{tab:ablation2}
    \begin{tabular}{l|c|c||c|c|c||c|c|c||c|c|c||c|c|c}
        \hline
         \multicolumn{15}{c}{Source macromoleculars: 1bxn, 1f1b and 1yg6}\\
         \hline
         N$_{sample}$ & mIoU & Dice & $\rho$ & mIoU & Dice & $\lambda_{1} \rightarrow \lambda_{4}$ & mIoU & Dice & $\sigma_{d}$ & mIoU & Dice & $\sigma_{r}$ & mIoU & Dice\\
         \hline
         5 & 41.2 & 57.1 & $8.4\%$ & 41.7 & 57.3 & [0.1, 0.1, 0.4, 0.4] & 44.4 & 60.2 & 100 & 49.3& 64.9& 0.8 & 47.5 & 63.2\\
         \textbf{10}& \textbf{50.3} & \textbf{65.9} & $17.8\%$ & 43.5 & 59.6 & \textbf{[0.2, 0.2, 0.3, 0.3]} & \textbf{50.3} & \textbf{65.9} & \textbf{120} & \textbf{50.3} & \textbf{65.9} & 1.0 & 48.0 & 63.8\\
         15 & 43.6 & 59.4 & $\textbf{24.4\%}$ & \textbf{50.3} & \textbf{65.9} & [0.3, 0.3, 0.2, 0.2] & 45.3 & 61.0 & 140 & 49.1 & 64.5 & \textbf{1.2} & \textbf{50.3} & \textbf{65.9} \\
         20 & 42.6 & 58.3 & $42.2\%$ & 41.0 & 56.8 & [0.4, 0.4, 0.1, 0.1] & 42.2 & 57.8 & 160 & 47.0 & 62.5 & 1.4 & 49.5 & 65.2\\
         \hline
    \end{tabular}
\end{table*}

\subsubsection{Effectiveness of the Improved Bilateral Filter}

We conduct a comprehensive set of experiments to validate the effectiveness of the proposed improved Bilateral Filter (IBF) for denoised pseudo-labeling. We conduct experiments using the three different denoisers introduced in Sec~\ref{sec:DPL} respectively, and report the segmentation results in both Table~\ref{tab:source1} and Table~\ref{tab:source2}. As can be seen from the tables, comparing the method of using NGM as a denoiser, employing Bilateral Filtering (BF) indeed brings a performance improvement (\ie, $mIoU$ increased $1.2\%$ in Table~\ref{tab:source1} and $Dice$ increased $1.0\%$ in Table~\ref{tab:source2}). This is because BF can preserve some edge information while denoising, thereby avoiding pseudo-label distortion. However, as discussed in Sec~\ref{sec:DPL}, range kernel of BF is not suitable for grayscale inputs. Our proposed IBF addresses this drawback by using a Laplacian transform, which allows the range kernel to focus more on gradient changes in the voxel space rather than value changes. Therefore, our new model achieves the best performance via using the proposed improved Bilateral Filter (\ie, $mIoU$ increased $2.4\%$ in Table~\ref{tab:source1} and $Dice$ increased $3.3\%$ in Table~\ref{tab:source2}).

\begin{figure}[!t]
    \centering
    \includegraphics[width=1\linewidth]{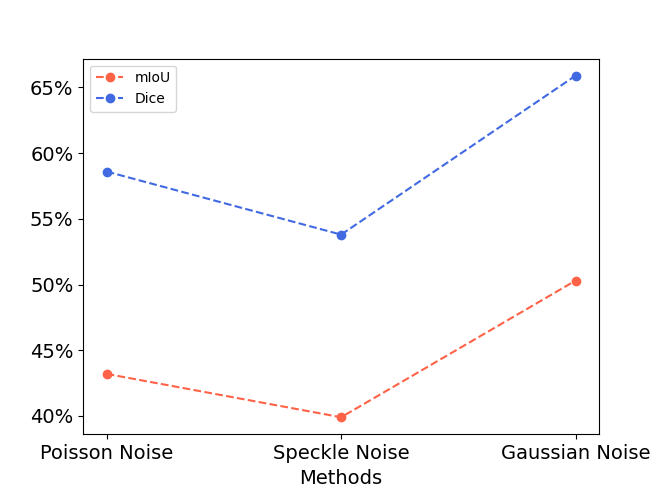}
    \caption{Ablation study of the proposed noise generation process. We set two different types of noise, Poisson noise and Speckle noise, to replace the Gaussian noise used in NGM for noise generation. }
    \label{fig:ablation}
\end{figure}

\subsubsection{Effectiveness of the Noise Generation Process}~\label{sec:NGP}
As mentioned in Sec~\ref{sec:NGM}, we choose Gaussian noise in NGM for noise generation. To demonstrate the rationality of our choice, we provide additional experiments in Fig~\ref{fig:ablation}, using Poisson noise~\cite{carlavan2011sparse} and Speckle noise~\cite{maity2015comparative} as the added noise for NGM, respectively. Given the variance $\sigma^{2}_{t}$ obtained through the average noise level $\overline{x}^{t'}_n$ of the subset, the Poisson noise can be represented as $\epsilon_{p} = \pi(\sigma_{t})$, and the Speckle noise can be formulated as
\begin{equation}
    \epsilon_{s} = x^{s}_{i} \times \mathcal{N}(0, \sigma_{t}^2 \mathbf{I}),
\end{equation}
where $x^{s}_{i}$ denotes the input source image. As can be seen from the figure, compared with the other two noises, using Gaussian noise achieves the best performance.

\subsubsection{Effectiveness of Different Proposed Modules}
We evaluate our Vox-UDA following the same experimental setting in Table~\ref{tab:source1} for the ablation study and use Vox-UDA (w IBF) as the final result of our proposed method. Table~\ref{tab:ablation1} shows the evaluation of the effectiveness of each module in our method. For comparison, we build a baseline only using a single discriminator with VoxResNet, which shows the same model structure as DANN~\cite{ganin2016domain}. From Table~\ref{tab:ablation1}, we observe that both ``w/o NGM'' and ``w/o PL'' achieve better performance than ``Baseline'', which demonstrates the effectiveness of our proposed two modules in dealing with the challenges for UDA in subtomogram segmentation. In the meantime, compared with using these two modules only, Vox-UDA achieves a significant performance improvement (\ie, $mIoU$ improved by $15.8\%$ and $Dice$ improved by $11.0\%$).

\subsubsection{Hyperparameter Analysis}
We herein further evaluate the hyperparameters in our approach. As shown in Table~\ref{tab:ablation2}, we evaluate the sampled number $N_{sample}$, the high-pass filter rate $\rho$, the weight $\lambda_{n}$ for consistency losses, the domain hyperparameter $\sigma_{d}$ and range hyperparameter $\sigma_{r}$. $N_{sample}$ is used to control the number of sampled targets for noise generation. As we discussed in the previous sections, the main goal of the noise generation module is to simulate target-like noises for the inputs from the source domain. Because the noise level of the whole target domain dataset is not evenly distributed, we choose to use random sampling instead of the whole dataset. Therefore, $N_{sample}$ is the key point. Either being too large or too small for such a number will lead to a negative impact on the model's performance. As the results reported in Table~\ref{tab:ablation2}, $N_{sample} = 10$ achieves the best performance (\ie, $mIoU$ increased $18\%$ compared to $N_{sample} = 20$ and $Dice$ increased $11\%$ compared to $N_{sample} = 15$). $\rho$ is the filter rate to control how much information is retained for further processes. As aforementioned, noise is usually contained in the high-frequency information of 2D or 3D images. However, the object information between the high frequency and the low frequency of the high-pass filter is typically determined by subjective discretion. Hence, we try different percentages of how much high-frequency information should remain to see which works better in our framework. Experimental results prove that $\rho = 24.4\%$ works the best in our proposed approach (\ie, $mIoU$ increased $16\%$ compared to $\rho = 17.8\%$ and $Dice$ increased $10\%$ compared to $\rho = 8.4\%$). $\lambda_{n}$ is the weight we set to control the relative importance among different consistency losses. As mentioned in Sec~\ref{sec:NGM} that for segmentation tasks, high-level features focus more on the edge details and low-level features focus on the textual information, we set the same value for $\lambda_{1}$ and $\lambda_{2}$ for low-level features, and the same value for $\lambda_{3}$ and $\lambda_{4}$ for the high-level ones (see Eq.\ref{eq:consis}). As can be seen in Table~\ref{tab:ablation2}, $\lambda_{1} = 0.2$, $\lambda_{2} = 0.2$, $\lambda_{3} = 0.3$ and $\lambda_{4} = 0.3$ achieve the best performance compared to other settings (\ie, our approach increases $19\%$ on $mIoU$ and $14\%$ on $Dice$). We further evaluate the effectiveness of the $\sigma_{d}$ and $\sigma_{r}$. Experimental results demonstrate that our settings are the most reasonable, increasing or altering $\sigma_{d}$ and $\sigma_{r}$ would not lead to an improvement in the model's performance. 

\section{Conclusion}\label{sec:conclu}
In this paper, we propose the first voxel-level unsupervised domain adaptation approach, termed Vox-UDA, for the subtomogram segmentation task. In detail, our Vox-UDA consists of a Noise Generation Module (NGM) and a denoised pseudo-labeling (DPL) strategy. NGM takes a subset of target samples as input and generates target-like Gaussian noise for the source domain data. DPL is based on a student-teacher learning framework, using a denoised target domain data to produce pseudo label for the original target data to boost the segmentation performance by reducing the effect of domain shift. Additionally, we propose an improved bilateral filter (IBF) to provide denoised target data for DPL, thereby enhancing the quality of the pseudo labels. The proposed IBF utilizes a 3D Laplacian operator to calculate the gradient of each voxel in the $h$, $w$, and $d$ directions, and replaces value differences with gradient differences to enhance the performance of bilateral filtering in the grayscale space. We have conducted large-scale experiments to demonstrate the prominent performance of our method. We anticipate our novel method can contribute more to the research in cryo-ET in terms of methodology and possibly enhanced  intepretability. Furthermore, we would propose that future research endeavors focus on enhancing the scalability of our method for a broader range of biomedical 3D image segmentation tasks.

\section*{Acknowledgments}
The authors acknowledge NVIDIA and its research support team for the help provided to conduct this work. This work was partially supported by the Australian Research Council (ARC) Industrial Transformation Training Centres (IITC) for Innovative Composites for the Future of Sustainable Mining Equipment under Grant IC220100028. This work was partially supported by U.S. NIH grants R01GM134020 and P41GM103712, NSF grants DBI-1949629, DBI-2238093, IIS-2007595, IIS-2211597, and MCB-2205148. This work was supported in part by Oracle Cloud credits and related resources provided by Oracle for Research, and the computational resources support from AMD HPC Fund.

{
\bibliographystyle{IEEEtran}
\bibliography{ref}
}

\end{document}